\documentclass{svproc}
\usepackage{url}
\usepackage{graphicx}
\usepackage{booktabs} 

\usepackage{CJKutf8}

\begin{document}
\mainmatter              
\title{Evaluating the Translation Performance of Large Language Models Based on Euas-20
}
\titlerunning{Evaluating the Translation Performance}  
%
\author{Yan Huang\inst{1} \and Wei Liu\inst{1}
}
\authorrunning{Yan Huang et al.} 
%
\tocauthor{Yan Huang, Wei Liu}
\institute{College of Software, Zhengzhou University of Light Industry,\\
\email{lwei230215@gmail.com}\\
}

\maketitle              

\begin{abstract}

In recent years, with the rapid development of deep learning technology, large language models (LLMs) such as BERT and GPT have achieved breakthrough results in natural language processing tasks. Machine translation (MT), as one of the core tasks of natural language processing, has also benefited from the development of large language models and achieved a qualitative leap. Despite the significant progress in translation performance achieved by large language models, machine translation still faces many challenges. Therefore, in this paper, we construct the dataset Euas-20 to evaluate the performance of large language models on translation tasks, the translation ability on different languages, and the effect of pre-training data on the translation ability of LLMs for researchers and developers. 
\end{abstract}
\section{Introduction}
The application and performance of Large Language Models (LLMs) on translation performance has become an important research direction and practical achievement in the field of modern natural language processing. In the recent emergence of large language models (LLMs), e.g., GPT-3 and GPT-4 , their translation performance on Zero-shot can be compared to that of powerful fully supervised machine translation systems\cite{jiao2023chatgpt,robinson2023chatgpt,moslem2023adaptive,zhu2024towards}. However, the massive corpus used for training big language models is usually dominated by monolingual data, in which the English corpus is dominant, while the proportion of corpus in other languages is relatively small \cite{zhu2023unsupervised,zhu2024mining}. 
Under this data distribution, whether the big language models can effectively model the correspondence between different languages and learn reliable translation knowledge is a great concern for researchers \cite{DBLP:journals/ipm/ZhuPX24}.
Models may face challenges in handling translation tasks between these languages. Therefore, we evaluate the popular large language models currently available in the market to acquire a better perception of the translation performance of large language models.

In this paper, the translation ability of Large Language Models (LLMs) is investigated by answering two questions, 1) What is the translation ability of LLMs?2) Factors affecting the translation ability of LLMs?3) What are the translation results of the LLMs?

In response to the first question, we evaluate several popular LLMs: English language-centric LLMs including Llama2\cite{touvron2023llama} , Falcon \cite{almazrouei2023falcon}, Vicuna\cite{zheng2024judging} , Mistral \cite{jiang2023mistral} and multilingual LLMs including Bloom and Bloomz\cite{le2023bloom} , Gemma\cite{team2024gemma} . In order to prevent data leakage and get more accurate results, we constructed a dataset Euas-20(A representative selection of 20 languages). LLMs translate other languages into Chinese and English respectively. The results show a significant improvement in the multilingual translation ability of LLMs. This improvement is not only reflected in the increase of the model's parameters, but also due to the model's improvement in training data and methods. We compare the translation results of LLMs on different languages. We find that there is a significant difference in the translation performance of LLMs on different languages, and LLMs perform better than Chinese when translating into English. For languages similar to English, LLMs also demonstrate better translation performance. Meanwhile, we find that LLMs also have translation ability on zero-resource languages. This suggests that the large language models have some generalisation ability and are able to establish correspondences between different languages in the absence of direct training data.

To address the second question, we analysed the LLMs by collecting information from their corpora. We find that a high-quality and diverse corpus can significantly improve the translation performance of LLMs. Multi-language and multi-domain training data can not only enhance the generalisation ability of the model, but also improve its effectiveness in different languages and different domains.

To address the third question, we have analysed the translation results of LLMs in various aspects.The translation results of LLMs are subject to some illusory problems, and their fluent output may mislead the users and make it difficult for them to identify the errors in the translation. In addition, LLMs tend to choose the most appropriate translation words in translation tasks by analysing a variety of factors such as semantics, fluency and culture. We also found that when LLMs met words that they had not encountered during model training, the models could not understand or process them accurately due to the lack of training on these words.

The purpose of this paper is to review and analyse the performance of the current large language model on translation tasks, to explore whether the large language model can effectively model the correspondence between different languages and the factors affecting translation, for the reference of researchers and developers.

\section{Background }
\subsection{Large Language Models}
Large Language Models (LLMs) have made significant progress in translation performance. Based on deep learning, especially the Transformer architecture\cite{vaswani2017attention,cui2024efficiently,zhu2023tjunlp} , these Large Language Models have learned rich linguistic knowledge by pre-training on a large amount of textual data, thus achieving excellent performance in various downstream tasks. 

The training process of a large language model is divided into two main phases. The first is the pre-training phase, in which the model learns unsupervised on large-scale textual data to master the basic structure and lexical usage of the language. The goal of this phase is for the model to learn a generic language representation. Next is the fine-tuning phase, in which the pre-trained model is subjected to supervised learning on a specific translation task using a bilingual parallel corpus to equip it with the ability to translate specific language pairs.

 The big language models support multiple languages, demonstrating the ability to generalize across languages. However, the training data of the big language models are dominated by the English corpus, with a smaller proportion of data in other languages, and this unbalanced data distribution poses a severe test for the performance of the models in a multilingual environment. Researchers are actively exploring ways to address these issues in order to further improve the performance of large language models in translation tasks.
\subsection{Machine Translation }
 Machine Translation (MT) is a technology that uses computers to automatically translate text from one language to another. In recent years, with the rapid development of artificial intelligence and natural language processing technology, especially the emergence of large language models (e.g., OpenAI's GPT series and Google's BERT), the ability of machine translation has been significantly improved.

Modern machine translation systems mainly rely on Neural Machine Translation (NMT) technology\cite{cho2014learning,bahdanau2014neural} . NMT utilises deep learning and neural network models, and is able to efficiently capture and process complex relationships between source and target languages through encoder-decoder architectures and self-attention mechanisms. Compared with traditional rule-based methods and statistical machine translation (SMT)\cite{zens2002phrase,koehn2007moses}, NMT performs better in terms of translation accuracy, fluency, and context understanding.

Machine translation, as one of the core tasks of natural language processing, has also benefited from the development of large language models and achieved a qualitative leap. However, machine translation still faces challenges, including translation of low-resource languages and maintaining coherence and fluency of translation in long texts \cite{koehn2017six}.

\begin{table}
\centering

\caption{Language}
\begin{tabular}{l l l l l}
 \toprule 
iso& Language& Language grouping& Script& Resource Level
\\
 \midrule 
ar& arabic& arabic& Arabic& Medium
\\
zh& chinese& Sino-Tibetan& Han& Medium
\\
da& danish& Germanic& Latin& Medium
\\
en& english& Germanic& Latin& High
\\
fr& french& Romance& Latin& High
\\
de& german& Germanic& Latin& High
\\
el& greek& Hellenic& Greek& Medium
\\
hi& hindi& Indo-Aryan& Devanagari& Medium
\\
is& icelandic& Germanic& Latin& Medium
\\
it& italian& Romance & Latin& High
\\
ja& japanese& Japonic& Kanji; Kana& Medium
\\
ko& korean& Koreanic& Hangul& Medium
\\
nl& nederlands& Indo-European& Latin& Medium
\\
no& norsk& Indo-European& Latin& Medium
\\
pt& portuguese& Romance& Latin& High
\\
ru& russian& Slavic& Cyrillic& High
\\
es& spanish& Romance& Latin& High
\\
tl& tagalog& Malayo-Polynesian& Latin& Low
\\
th& thai& Sino-Tibetan+Kra-Dai& Thai& Medium
\\
vi& vietnamese& Vietic& Latin& Medium
\\
 \bottomrule 

\end{tabular}
\end{table}

\section{Experimental Setup}
\subsection{Dataset }
In order to evaluate the real translation capabilities of large language models, we constructed a dataset called Euas-20. This dataset contains twenty representative languages (Table 1), covering a large part of the global population, while demonstrating a diverse background of writing systems and language families. We have selected a number of important languages that not only have a large number of speakers, but also include some languages that are considered under-resourced in the research community. With this diverse dataset, we are able to comprehensively evaluate the translation performance of the large language models in different language contexts, and thus gain a more accurate understanding of their performance in real-world applications.

\begin{figure}
    \centering
    \includegraphics[width=0.5\linewidth]{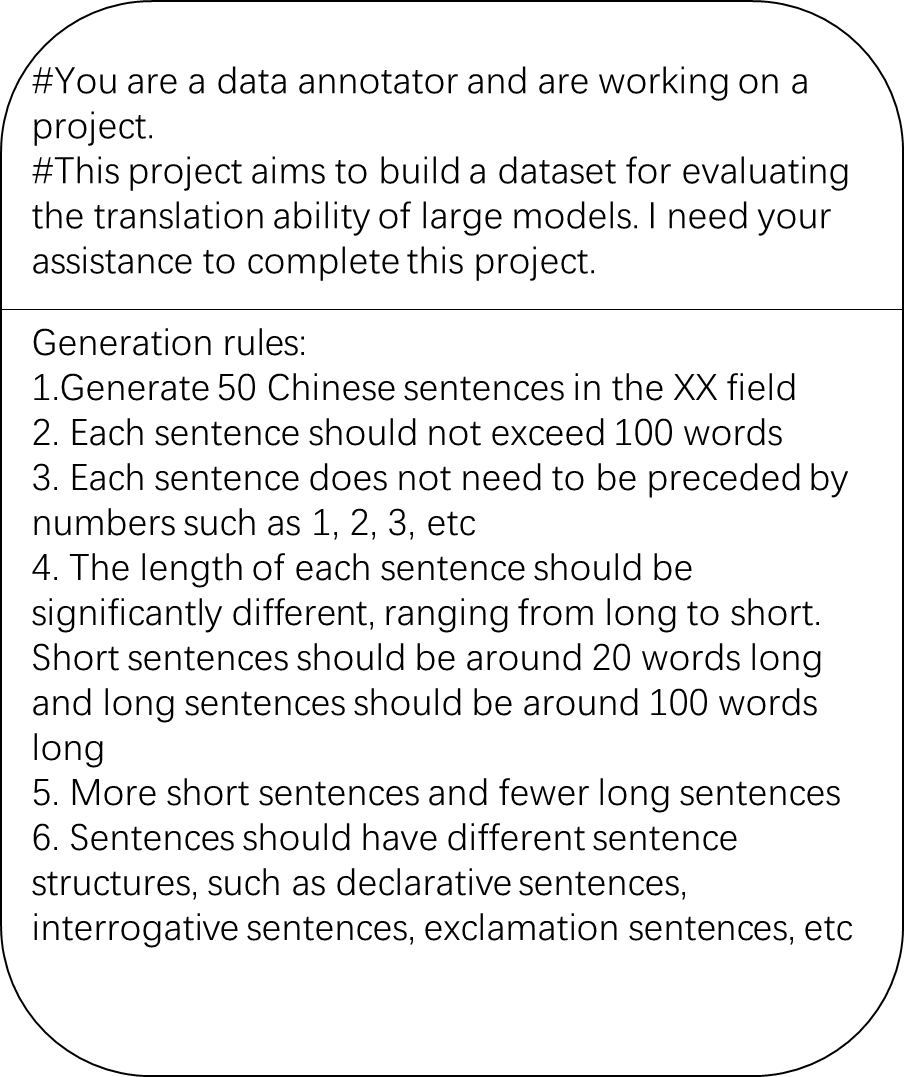}
    \caption{Prompt 1}
    \label{fig:1l}
\end{figure}

languages we use in our work are listed in Table 1. Referring to the information provided in Goyal et al. (2022)\cite{goyal2022flores}, we populated the table. For each language, we show the ISO code, language name, language grouping, alphabet and resource level.

We selected about twenty domains such as medicine, science, art, education, environment, finance, entertainment, sports, politics, agriculture, etc. to ensure a wider coverage of the dataset. After that, we designed a prompt (Fig.\ref{fig:1l}) and fed it into ChatGPT, allowing it to act as a data annotator and generate sentences according to specified rules. In each domain, ChatGPT generated fifty sentences, including different sentence types such as declarative, interrogative and exclamatory sentences. We deleted sentences with high similarity and repeated the process, eventually selecting about fifty different sentences in each domain and constructing a document containing one thousand Chinese sentences.

Next, we used Google Translate to translate this Chinese document into other target languages to build a complete dataset. In this way, we ensure that the dataset is diverse and representative, and we are able to more comprehensively evaluate the translation capabilities of large language models across different domains and languages.

\subsection{ LLMs }
We evaluated the translation capabilities of nine currently popular LLMSs: falcon7b, mistral-7b, Llama-2-7b-hf, bloom-7b1, bloomz-7b1-mt, Meta-Llama-3-8B, mpt-7b, vicuna-7b, and gemma-7b.

\begin{figure}
    \centering
    \includegraphics[width=0.5\linewidth]{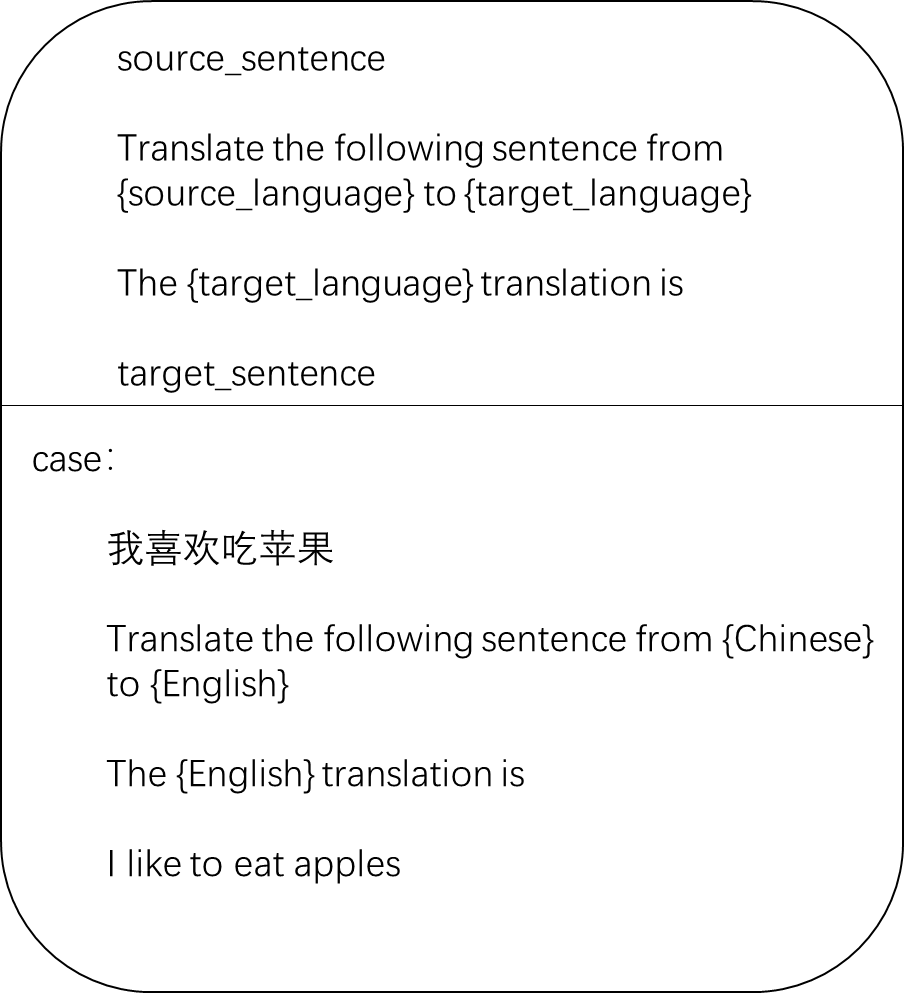}
    \caption{Prompt 2}
    \label{fig:2l}
\end{figure}

\subsection{ Evaluation Methods}

Focusing on Chinese and English, through a prompt (Fig. \ref{fig:2l}), 'source-sentence' stands for the original sentence and 'target-sentence' stands for the target sentence, and the original sentence is input to the LLMs by the command (Translate the following sentence from 'source-language' to 'target-language' and The 'target- language' translation is), so that the LLMs can translate and output the target sentence under Zero-Shot learning. In this way, various languages in the dataset are translated into Chinese and English.

\subsection{ Evaluation Indicators}

Evaluation metrics are an important measure of translation quality. We adopt commonly used automatic evaluation metrics including BLEU\cite{papineni2002bleu}, which calculates translation accuracy by comparing the n-gram overlap between candidate and reference translations, which is the traditional method for assessing the quality of machine translation.

In addition, we also consider the emerging metric COMET\cite{rei2020cometneuralframeworkmt}, which is designed to learn to predict human judgements of machine translation quality, and which better reflects subjective human assessments of translation quality. By combining these evaluation metrics, we are able to assess the translation performance of large-scale language models in a more comprehensive way, ensuring the accuracy and reliability of the assessment results.

\section{Testing of machine translation for LLMs}

\begin{figure}
    \centering
    \includegraphics[width=1\linewidth]{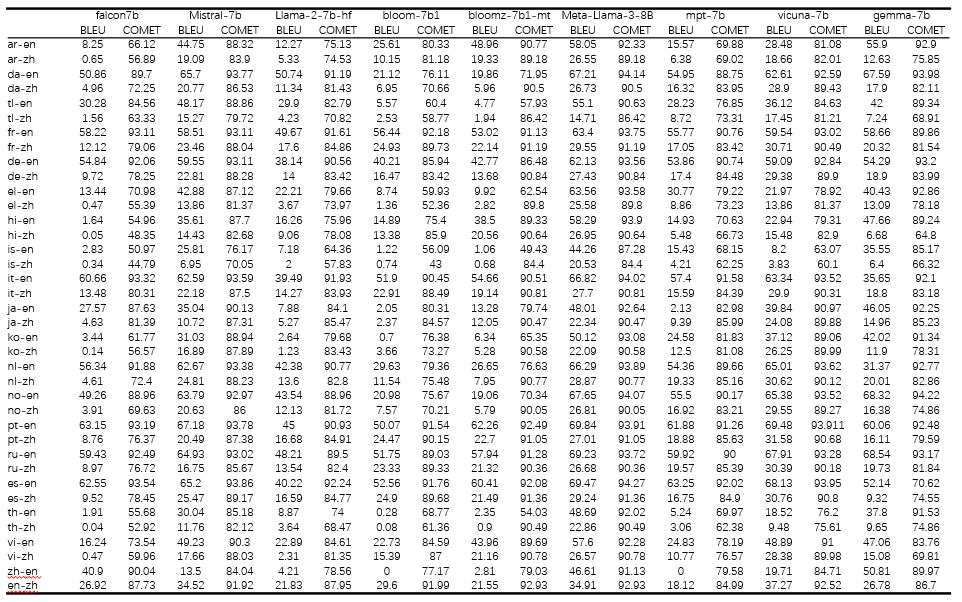}
    \caption{BLEU and COMET scores for nine LLMs translations centred on English and Chinese.}
    \label{fig:3l}
\end{figure}

In this section, we report the results of the translation of LLMs (Fig.\ref{fig:3l}) and analyse the translation performance of LLMS.

\subsection{Continuous Improvement of Translation Ability of LLMs}

\begin{figure}
    \centering
    \includegraphics[width=1\linewidth]{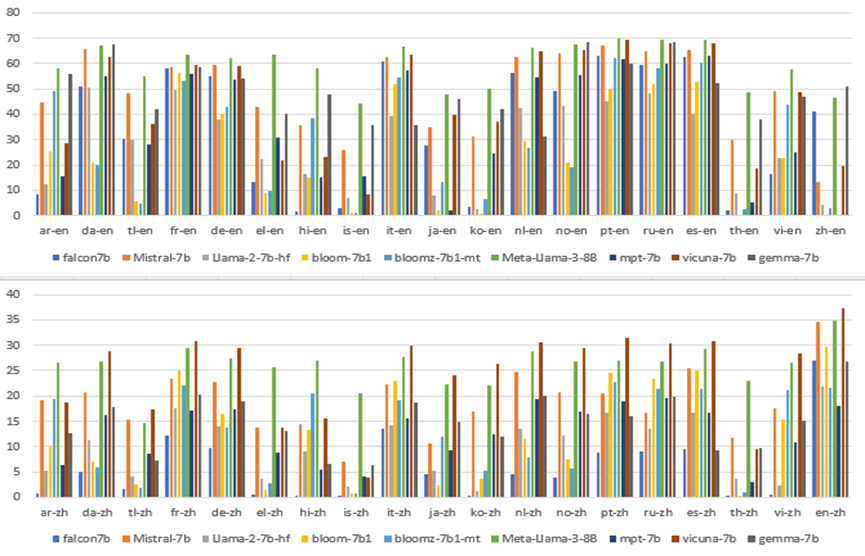}
    \caption{Translation performance (BLEU) of LLMS on our evaluated languages, ‘xx-en’ and ‘xx-zh’ denote translation from other languages to English and Chinese, respectively. }
    \label{fig:4l}
\end{figure}

In recent years, the multilingual translation capability of Large Language Models (LLMs) has been significantly improved. Even under Zero-Sample Learning (Zero-Shot) conditions, LLMs still exhibit good translation performance in most translation directions, as shown in Fig. \ref{fig:4l}. Based on the  scores of LLMs on different languages  , we can find that the translation ability of LLMs has gradually improved, especially the recent LLMs have reached new heights in terms of translation performance. For example, Llama-3-8B significantly outperforms the previous Llama-2-7B, and vicuna-7B outperforms Llama-2-7B. Overall, Llama-3-8B performs the best among all the LLMs evaluated, and it obtains the highest BLEU and COMET scores in most translation directions, showing its superior translation capabilities.

This progress is not only reflected in the increase of the model's parameters, but also due to the model's improvement in training data and methodology. Llama-3-8B uses larger and higher quality multilingual datasets during training, and adopts more advanced training algorithms, which enable it to maintain a high level of translation quality even when dealing with complex and rare language pairs. At the same time, the model's architectural optimisation and inference speed have also been improved, making Llama-3-8B not only more accurate but also more efficient in practical applications.

In addition, Llama-3-8B and other advanced LLMs demonstrate a high degree of flexibility and adaptability in coping with multilingual text comprehension and generation tasks. These models can play an important role in cross-cultural and cross-linguistic communication.

\subsection{Translation performance of LLMSs across languages}

The translation performance of large-scale language models (LLMs) varies significantly across languages. Typically, LLMs translate well on high-resource languages, but have relatively poor translation performance on low- and medium-resource languages. We find that LLMs perform particularly well when translating into English and relatively poorly when translating into Chinese. For languages similar to English, LLMs also demonstrate better translation performance. For example, LLMs generally achieve better translation results in the Indo-European Romance and Germanic languages.

For some languages that are more different from English, such as the Tai-Kadai languages, LLMs produce very poor translation results. This uneven translation performance is mainly due to the differences in the training data, where the high volume and quality of data for high-resource languages make the models perform better on these languages. On the other hand, low and medium resource languages are difficult for the model to learn enough linguistic features due to the scarcity of data, resulting in unsatisfactory translation results.

Nevertheless, LLMs show some translation ability even on zero-resource languages. This suggests that the large language models possess some generalisation ability and are able to establish correspondences between different languages in the absence of direct training data. Behind this ability is the fact that the models have learnt common features and structures between languages through large-scale multilingual training, so that they can still translate reasonably well in the face of new language pairs.

\subsection{Effect of corpus on the translation performance of LLMs}

By analysing pre-training data and corpora of large-scale language models (LLMs), we can investigate the relationship between translation performance and corpus size and category. By collecting LLMs with training data sizes (Table \ref{Table 2}), we find that the larger the pre-training data size, the better the translation performance of the LLMs.For example, Llama-3-8B and Gemma-7B outperform other models overall. This suggests that rich training data is one of the key factors to improve the translation ability of the models.

\begin{table}
\centering
\caption{Training volume of LLMs}
\label{Table 2}
\begin{tabular}{l l}
\hline
LLM & Token \\
\hline
Gemma 7b & 6 trillion \\
Llama-2-7b-hf & 2 trillion \\
mpt-7b & 1 trillion \\
Meta-Llama-3-8B & 15 trillion \\
\hline

\end{tabular}

\end{table}

Most of the pre-training corpora of the big language models are English-centric, which on the one hand makes the models perform extremely well in English language ability; on the other hand, it also leads to their weaker ability in non-English languages. This English-centric corpus configuration improves the efficiency of the model in handling English-related tasks, but the model's performance appears to be insufficient when dealing with other languages, especially low- and medium-resource languages.
\begin{figure}
    \centering
    \includegraphics[width=0.5\linewidth]{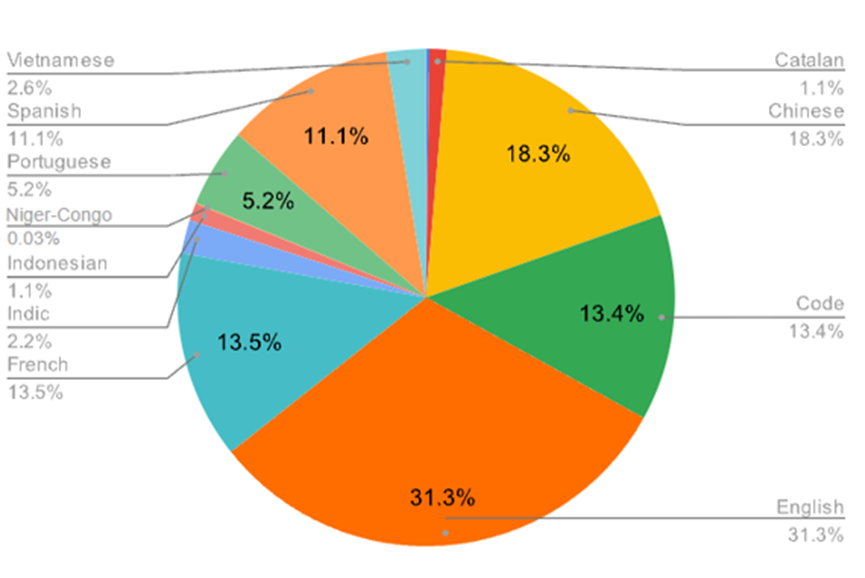}
    \caption{Corpus share of LLMs}
    \label{fig:5l}
\end{figure}

Models trained in multiple languages have achieved better results in translation than LLMs limited to one or a few languages. For example, the translation performance of Gemma-7B is significantly better than that of Falcon-7B.Meanwhile, when translating languages, multilingual models tend to have better translation ability for languages that have been pre-trained than for languages that have not been pre-trained. For example, the corpus share of bloom-7b1 and bloomz-7b1-mt (Fig. \ref{fig:5l}) has better translation ability for pre-trained languages. This suggests that multi-language training can effectively enhance the model's translation capability, enabling it to better handle translation tasks between various language pairs.

From these observations, it can be found that a high-quality and diverse corpus can significantly improve the translation performance of LLMs. Multi-language and multi-domain training data can not only enhance the generalisation ability of the model, but also improve its effectiveness in different languages and different domains. Therefore, in order to meet the translation needs of various languages, future LLMs should make full use of diverse corpora in the pre-training process and continuously increase the proportion of data from low- and medium-resource languages. 

\subsection{Illusions in the translation of LLMs}
Neural Machine Translation (NMT) is a task that translates a source language into a target language through inference and relies on parallel data samples used for training. Compared to Statistical Machine Translation (SMT), the output of NMT is usually very fluent, with a quality close to the human level. However, this poses a potential problem: when NMT hallucinates (i.e., generates inaccurate or spurious translations), its smooth output may mislead users and make it difficult for them to identify errors in the translation.

\begin{CJK}{UTF8}{gbsn}
By analysing the translation results of LLMs, we classified the hallucinations in NMT as two categories \cite{zhou2021detectinghallucinatedcontentconditional}, intrinsic and extrinsic hallucinations. Intrinsic Illusion: Incorrect information is included in the translation that does not match what is in the source text. An example of such an illusion is ‘不太了解 ’, which negates ‘了解多少 ’ in the source text. Extrinsic illusions: the translation produces additional content that does not exist in the source text.’我忘了带手机 ’ is an example of illusory content because it is added without any apparent connection to the input.
\end{CJK}

\begin{table}
\centering
\caption{Illusions in the translation of LLMs}
\label{Table 3}
\begin{CJK}{UTF8}{gbsn}
\begin{tabular}{l l l l}
\hline
Category & Source & Correct Translation & Hallucinatory Translation \\
\hline
Intrinsic & Excuse me, how much do you  & 请问，你对这项技 & 非常抱歉，我们这么说是 \\
  & know about this technology? & 术的了解有多少？ & 因为我们不太了解这种技术。 \\
Extrinsic & Excuse me, how much do you& 请问，你对这项技 & 对不起，我忘了带手机了。 \\
  & know about this technology? & 术的了解有多少？ &   \\
\hline

\end{tabular}
\end{CJK}
\end{table}

The results show that as the pre-trained corpus continues to grow, the pre-trained model becomes more and more effective in generating faithful summaries of human assessments. By comparing the translation results of Gemma-7B and Falcon-7B, more intrinsic illusions are generated for monolingual models; while for multilingual models, this is less frequent. Also, we found that nouns are the most hallucinated words, and verbs also account for a certain number of hallucinations. In addition, LLMS tend to be more prone to intrinsic hallucinations when confronted with untrained language.

Therefore, it is crucial to improve the accuracy and reliability of machine translation. By continuously improving our models, enhancing the diversity and quality of our datasets, and using more advanced evaluation metrics to detect and reduce illusions in translation, we can mitigate these risks and provide more secure and reliable translation services.

\subsection{Translation words that LLMs tend to choose in translation tasks}
\begin{CJK}{UTF8}{gbsn}
This section explores the translation words that Large Language Models tend to choose in translation tasks and the reasons behind them.

Through previous analyses of the translation results of Gemma-7B and Falcon-7B, we found that LLMs tend to choose common word collocations in the target language during the translation process. This not only improves the naturalness of the translation, but also makes it more in line with the usage habits of the target language. For example, ‘make a decision’ in English is often translated as ‘做决定’ instead of ‘制造决定’ in Chinese because the former is a common collocation in Chinese and is more in line with the language conventions. This is because the former is a common collocation in Chinese, which is more in line with the language convention.

In addition, we also found that the model selects those words that are closest in meaning to be translated by deep understanding of the original and the translated text. For example, when translating the English word ‘computer’ into Chinese, the model chooses ‘电脑’ instead of ‘计算机’ because ‘电脑’ is a common collocation in modern Chinese. ‘is more commonly used and semantically accurate in modern Chinese.

LLMs tend to choose the most appropriate translation words in translation tasks by comprehensively analysing various factors such as semantics, fluency and culture. This approach not only improves the accuracy and naturalness of the translation, but also makes the translation result more in line with the usage habits of the target language.
\end{CJK}
\subsection{Phenomenon of unregistered words}

Out-of-vocabulary words (OOV words), refer to words that have not been encountered during model training. These words may be new terms, technical terms, foreign language vocabulary, or recently emerged buzzwords. We found that due to the lack of training on these words, the model cannot understand or process them accurately. We choose the translation results of Gemma-7B and Falcon-7B as representative.

For monolingual models, when the model is confronted with words that have not been trained across languages, such as ‘madilim na bagay’ (dark matter) in Filipino, the model will ignore or mistranslate them to other nouns. For multilingual models, even if the model has been trained cross-linguistically, when the model encounters a new word like ‘schadenfreude’ (a German word that refers to the emotion of taking pleasure in someone else's misfortunes), it may not be able to correctly understand the meaning because the word has never appeared in its training data. ever appeared in its training data. As a result, the model will choose to ignore it, not translate it or translate it incorrectly.

In the future, LLMs need to increase their training data to cover a wider range of vocabulary, as well as dynamically expand the model's vocabulary by using external resources such as vocabularies or online data sources; to reduce this phenomenon and to improve the translation ability of LLMs.

\section{Related Work}

In the field of large language model translation capability evaluation, there have been a large number of related studies devoted to exploring the translation performance of different models on multiple language pairs and text types.Bang et al. (2023) \cite{bang2023multitask}and Hendy et al. (2023)\cite{hendy2023good} evaluated ChatGPT on 13 and 18 languages, respectively; Zhu et al. (2023)\cite{zhu2023multilingual} evaluated the translation capability of four popular large language models, XGLM, BLOOMZ, OPT, and ChatGPT, on 102 languages, on 202 directions. 202 directions The multilingual translation capabilities of four popular large language models, XGLM, BLOOMZ, OPT and ChatGPT, were evaluated.

In this paper, 20 representative languages are selected to evaluate nine current mainstream large-scale language models. The evaluation focuses on Chinese and English, but covers a wide range of other languages as well. Multilingual translations are performed with these models and the results are compared with a state-of-the-art translation engine (Google Translate) in order to comprehensively evaluate the translation capabilities and performance of these large language models. The aim of the study is to determine the performance of these models in different linguistic contexts, as well as their usability and accuracy in real translation tasks.

Despite the significant progress made by large-scale language models on translation tasks, a number of challenges remain. For example, there is still room for improvement in the model's ability to handle low-resource languages and diverse texts. Future research directions include improving the evaluation metrics, optimising the model structure and enhancing the training methods to further improve the performance and generalisation of large language models on translation tasks. These improvements will not only help to enhance the model's translation accuracy, but also enhance its adaptability in dealing with complex and diverse language environments.

\section{Conclusion}

In this paper, a dataset called Euas-20 is constructed and nine popular large language models (LLMs) are evaluated using this dataset. The evaluation process focuses on Chinese and English, and compares the translation performance of these models and their translation capabilities on various languages through translations in 20 languages. Also, the paper analyses the impact of pre-training data and corpus on the translation performance of large language models.The translation results of the LLMs were analysed in various ways.

The results show that although the translation performance of LLMs is improving, with Llama-3 performing particularly well, far exceeding other models, the translation ability of these models on different languages is still very unbalanced. Especially when dealing with low-resource languages, they still face great challenges. In addition, a high-quality and diverse corpus plays a significant role in improving the translation performance of large language models.

Future research needs to continue to explore how to enhance the translation capabilities of LLMs on more languages to achieve more balanced and comprehensive translation performance. This includes improving the model structure, optimising training methods, and extending and enhancing the quality and diversity of the corpus.

\bibliographystyle{References}

\end{document}